\documentclass[11pt]{article} 
\usepackage{rldmsubmit,palatino}
\usepackage{graphicx}
\usepackage{hyperref}
\usepackage{booktabs}
\usepackage{tabularx}
\usepackage{multirow}
\usepackage{amsmath}
\usepackage[numbers,sort&compress]{natbib} 
\title{Less is more? Rewards in RL for Cyber Defence}

\author{
Elizabeth Bates\\
The Alan Turing Institute\\
London\\
\texttt{ebates@turing.ac.uk}
\And
Chris Hicks\\
The Alan Turing Institute\\
London\\
\texttt{c.hicks@turing.ac.uk} \\
\And
Vasilios Mavroudis\\
The Alan Turing Institute\\
London\\
\texttt{vmavroudis@turing.ac.uk} \\
}

\begin{document}

\maketitle

\begin{abstract}


The last few years have seen an explosion of interest in autonomous cyber defence agents based on deep reinforcement learning. Such agents are typically trained in a cyber gym environment, also known as a cyber simulator, at least 32 of which have already been built. Most, if not all cyber gyms provide dense ``scaffolded'' reward functions which combine many penalties or incentives for a range of (un)desirable states and costly actions. Whilst dense rewards help alleviate the challenge of exploring complex environments, yielding seemingly effective strategies from relatively few environment steps; they are also known to bias the solutions an agent can find, potentially towards suboptimal solutions. This is especially a problem in complex cyber environments where policy weaknesses may not be noticed until exploited by an adversary. Sparse rewards could offer preferable or more effective solutions and have been overlooked by cyber gyms to date. In this work we set out to evaluate whether sparse reward functions might enable training more effective cyber defence agents. Towards this goal we first break down several evaluation limitations in existing work by proposing a ground truth evaluation score that goes beyond the standard RL paradigm used to train and evaluate agents. By adapting a well-established cyber gym to accommodate our methodology and ground truth score, we propose and evaluate two sparse reward mechanisms and compare them with a typical dense reward. Our evaluation considers a range of network sizes, from 2 to 50 nodes, and both reactive and proactive defensive actions. Our results show that sparse rewards, particularly positive reinforcement for an uncompromised network state, enable the training of more effective cyber defence agents. Furthermore, we show that sparse rewards provide more stable training than dense rewards, and that both effectiveness and training stability are robust to a variety of cyber environment considerations.

\acknowledgements{Research funded by the Defence Science and Technology Laboratory (Dstl) which is an executive agency of the UK Ministry of Defence providing world class expertise and delivering cutting-edge science and technology for the benefit of the nation and allies. The research supports the Autonomous Resilient Cyber Defence (ARCD) project within the Dstl Cyber Defence Enhancement programme.}

\end{abstract}


\startmain 

\section{Introduction}
Cyber attacks are increasingly frequent and sophisticated, straining already limited cyber defence resources worldwide. Consequently, there has been a rising level of interest in deep reinforcement learning (DRL) based autonomous cyber defence (ACD) agents which could provide both novel defensive strategies and automation for tasks that currently occupy human analysts. DRL is promising for this task because of its ability to learn complex policies from interactions and discover strategies unconstrained by potentially flawed security models.


Cyber gyms provide an efficient and controlled environment for ACD agents. This is particularly important for network security tasks, enabling the large number of interactions required for training without risking production networks or systems. Accordingly, many cyber gyms have been created to enable training agents that defend networked systems. Necessarily, each cyber gym defines one or more Markov Decision Processes (MDPs) in terms of a state space comprising network information, an action space of defensive activities, and a reward function that aligns with a defensive goal. 

The typical cyber gym reward function is densely scaffolded, combining multiple penalties and incentives which are issued for a variety of defensive actions and network states~\cite{yawningtitan, cage_cyborg_2023, bates2023reward}. Dense rewards may be preferable because they expedite learning, providing apparently effective solutions using fewer environment steps during training, but they also risk constraining agents to sub-optimal solutions~\cite{riedmiller18a}. This is especially concerning for ACD agents which might then contain avoidable weaknesses that are difficult to identify in advance of an attack. Furthermore, dense rewards draw numerical equivalences between diverse network states and actions. As the scale and complexity of the cyber task increase, this becomes increasingly impossible to manage properly and exacerbates the risk of undesirable agent behaviour.

At the expense of typically requiring more training iterations, sparse rewards place fewer constraints on the solution space and might enable preferable or more effective policies to be discovered. Existing work has not investigated the possibility that dense rewards might limit the performance of ACD agents trained using DRL. To investigate this risk, we (1) propose a ground truth scoring mechanism that allows for comparison between agents trained using different reward functions, (2) adapt a popular cyber gym to enable ground truth scoring between two sparse reward functions and a conventional dense reward, and (3) show that a sparse positive reward function can enhance the performance and training stability of ACD agents across a variety of network sizes, action types, and MDP models. 




\section{Methodology}



\subsection{Yawning Titan} 


Yawning Titan (YT) is a well-established cyber gym providing an abstract, graph-based network simulation environment for training defensive ACD agents ~\cite{yawningtitan}. YT is highly configurable, allowing for customisable game dynamics, actions and reward functions. We configure YT to simulate a linear network structure with a single fixed entry node for the red agent. The attacker can only spread from an already compromised node, moving laterally through the network. The goal of the blue agent is to defend the network by minimising the number of compromised nodes. Conversely, the red agent aims to compromise as many nodes as possible. 

The observation space comprises a vector which embeds the network adjacency matrix, the vulnerability of each node, and whether any nodes are isolated or compromised. We do not use node isolation, and the vulnerability of each node is set to $1$ which assumes a very powerful attacker whose attacks on a node will never fail. We create two action spaces: (1) the ``basic action space'' with only two actions: ``scan network'' and ``restore node'', and (2) the ``extended action space'' which also adds ``place decoy''. The place decoy action is a proactive defence replicating the use of a deceptive ``canary'', a technique sometimes used to detect and delay attackers in real world networks. The red action space includes two actions: ``do nothing'' and ``basic attack''.








\subsection{Ground Truth Evaluation} \label{agent score eval}

To the best of our knowledge, previous work on ACD is limited to evaluating agent performance only in terms of the average episodic reward over a large number of episodes. In other words, the MDP is presumed to model the ground truth and the episodic reward to accurately represent an agent's performance. However, cyber gyms are complicated environments that simulate both offensive (red) and defensive (blue) agent actions. According to the MDP framework, actions are taken during time steps which necessitates assigning an order in which the actions are taken. 

A problem with this method of evaluation is that episodic reward, calculated from the state at the end of each step, can omit network events that occur intra-step but are resolved before the reward is determined. Concretely--red agents may compromise nodes during the step, shortly before the blue agent removes the compromise, and this will not be reflected in the reward. During evaluation this makes it impossible to distinguish between some steps in which nodes were compromised and those in which no compromise occurred. To overcome this issue we introduce a ground truth score calculated from the state after every action, rather than every time step. We calculate this as a penalty of $-1$ for any time step in which a node is compromised, irrespective of the end-of-step state. This metric is applied to all the agents trained in our experiments, offering a score that can be usefully compared across agents trained using different reward functions. 
The ground truth score is a more accurate measure of how successfully an agent achieves the real objective of an ACD agent i.e., an uncompromised network.




\subsection{Evaluating Reliability} \label{agent reliability}

    
    It is important to consider the reliability of training a model, especially in the case of DRL where training performance is often unstable and phenomena such as catastrophic forgetting are not uncommon.
    To this end, we utilise a metric to evaluate the impact of reward function on training reliability. 
    
    Chan et al. (2020) propose dispersion variability (DV) across time as a metric of RL training reliability ~\cite{Chan2020Measuring}. DV calculates the inter-quartile range (IQR) of de-trended training data, isolating high-frequency performance fluctuations while excluding the variability from desirable policy improvement trends. Although Chan et al. describe several other reliability metrics, they are designed to assess variability across multiple training runs which is less applicable as we aggregate the results from multiple runs in our training curves. We calculate the  DV across time metric with de-trending to our average training curves over the 25 runs (hence $\bar{\text{DV}}$)  and present the results in Section~3.

\subsection{Experiments}

    We evaluate the performance and training stability of 3 different reward functions shown in Table \ref{tab:RewFuncDefinitions}: (1) a sparse positive reward that incentivises only a network entirely free of compromised nodes, (2) a sparse negative reward issuing a penalty any time the network is entirely compromised, and (3) a conventional ~\cite{cage_cyborg_2023, yawningtitan} dense reward function which issues a negative reward for each host that is compromised at the end of an environment step. By avoiding drawing any numerical equality or comparison between different nodes, which in practice would likely have different value to an attacker, the sparse rewards place fewer constraints on the optimisation objective. Correspondingly, we anticipate that the sparse reward functions could enable agents to learn more effective policies. 
\vspace{-2ex}
      \begin{table}[ht]
      \caption{Reward functions and their definitions.}
      \vspace{0.75ex}
    \centering
    \small
    \begin{tabular}{p{0.15\textwidth} p{0.38\textwidth}} 
    \toprule
    \textbf{Reward Function} & \textbf{} \\ 
    \midrule
    Sparse Positive & +1 when 0 nodes are compromised per time step \\
    Sparse Negative & -1 when all nodes are compromised per time step \\
    Dense & -1 per node compromised per time step \\
    \bottomrule
    \end{tabular}
    \label{tab:RewFuncDefinitions}
    \end{table}

    \vspace{-1.5ex}
    Informed by the insights provided by our ground truth scoring mechanism, we trained agents in using two possible orderings of red and blue agents actions: blue then red, and red then blue. We performed these experiments to understand whether any of our reward functions might train more effective cyber defence agents whilst also being robust to the order in which agents take turns during each step.

    
    Prior work using the "CAGE 2 challenge" cyber gym~\cite{kiely2023autonomous}, which features 12 nodes and a comparatively expansive action space, has shown that environment complexity and the inability to interpret agent behaviour scales rapidly as the network grows. Complex network simulations quickly obfuscate the relationship between reward function and outcomes in agent behaviour. With this in mind, we started our experiments with the smallest network sub-problem: 2 nodes and 2 actions for the red and blue agents. We then scaled the network gradually up to 50 nodes, and added the extended action space including the proactive decoy action. This iterative approach enabled us to evaluate how performance and training stability is impacted by the reward function as both the network size and action space increased.

    In all of our experiments the episode length is fixed at 100 steps, and each agent is trained using PPO--the DRL algorithm most commonly used for training ACD agents. The evaluation scores use the ground truth metric outlined in Section~\ref{agent score eval}, and reliability is evaluated using the average DV metric over total runs from Section~\ref{agent reliability}. All agents are trained using 25 different seeds for between 1 and 2.5 million time steps depending on network size and each resulting policy is evaluated for 1000 episodes. The average evaluation score from each policy is averaged over the 25 training runs, resulting in a final evaluation score for each network size, reward function, and agent-order we investigate.



    


\section{Results and Discussion}


    Firstly, we examine the impact of each reward function using the basic action space and the standard agent order of red then blue. We configure the red agent with a 90\% probability of performing a basic attack (which always succeeds) each time step, alternatively laying dormant i.e., does nothing 10\% of the time. Consequently, the best possible evaluation score for any defensive agent is $-0.9$. This would mean the red agent is making no progress through the network and is confined to, at most, a single node. Concerning $\bar{\text{DV}}$ across time (i.e., training reliability) a score of 0 is the best-case and indicates minimal fluctuations in the de-trended performance throughout training.
    As shown in Table \ref{tab:Simple action space}, only the sparse positive reward achieves an optimal average evaluation score for all network sizes. The negative and dense reward functions show significantly worse evaluation scores and also decrease in performance as the network size increases. The dense reward achieves an optimal score only in the smallest network size of 2 nodes. Considering variability during training, neither the positive nor the negative sparse rewards show significant de-trended performance fluctuations throughout training. Taken together, the positive sparse reward alone shows the combination of both high training reliability and optimal final performance.



    
\vspace{-0ex}

\begin{table}[ht]
\caption{Results for agents trained using 3 different reward functions for both agent orders in the basic action space.}
\vspace{0.1in}
\small
\centering
\renewcommand{\arraystretch}{1.2} 
\begin{tabular}{l c c c c c c c}
    \toprule
    \multirow{2}{*}{\textbf{Agent Order}} & \multirow{2}{*}{\textbf{Network Size}} 
    & \multicolumn{6}{c}{\textbf{Reward Function Evaluation -- Action space: Basic}} \\  
    \cmidrule(lr){3-8} 
    
    & & \multicolumn{2}{c}{\textbf{Sparse Postive}} & \multicolumn{2}{c}{\textbf{Sparse Negative}} & \multicolumn{2}{c}{\textbf{Dense}} \\  
    \cmidrule(lr){3-4} \cmidrule(lr){5-6} \cmidrule(lr){7-8}
    
    & & \textbf{Eval Score}  & \textbf{$\bar{\text{DV}}$ (e-3)} 
    & \textbf{Eval Score}  & \textbf{$\bar{\text{DV}}$ (e-3)} 
    & \textbf{Eval Score}  & \textbf{$\bar{\text{DV}}$ (e-3)} \\
    
    \midrule
    
    \multirow{5}{*}{\textbf{Red then Blue}}  
    & 2  & -0.90  & 0   & -1.37    & 0   & -0.90   & 0       \\
    & 5  & -0.90  & 0   & -3.78    & 0   & -0.98   & 0.01    \\
    & 10 & -0.90  & 0   & -6.15    & 0   & -1.57   & 0.66    \\
    & 20 & -0.90  & 0   & -9.07    & 0   & -4.99   & 4.27    \\
    & 50 & -0.90  & 0   & -19.72   & 0   & -7.68   & 14.97   \\ 
    
    \midrule
    
    \multirow{5}{*}{\textbf{Blue then Red}}  
    & 2  & -0.90    & 0.14   & -0.94     & 0.14    & -1.22   & 0.19    \\
    & 5  & -1.06    & 0.13   & -3.59     & 0.13    & -1.33   & 0.26    \\
    & 10 & -3.91    & 0.05   & -6.10     & 0.25    & -1.76   & 1.46    \\
    & 20 & -6.98    & 0.03   & -10.64    & 0.10    & -2.88   & 5.94    \\
    & 50 & -12.68   & 0.02   & -26.48    & 0       & -8.30   & 12.10   \\ 
     
    \bottomrule
\end{tabular} \label{tab:Simple action space}
\end{table}

Next, we consider the impact of agent order by maintaining the basic action space but with the alternative agent order of blue then red. As the red agent attack rate is unchanged from 90\%, the optimal evaluation score remains $-0.9$ on average and the best-case $\bar{\text{DV}}$ is still $0$. The results in Table \ref{tab: decoy action space} show that this setting increases the difficulty of network defence irrespective of the reward function. Only the sparse positive reward achieves the optimal average evaluation score for the smallest network size of 2 nodes. For networks with 5 nodes, the sparse positive reward also provides the best performance. Thereafter, performance decreases and the dense reward shows superior, although sub-optimal, average scores. Compared with the red then blue agent order, $\bar{\text{DV}}$ across time scores are greatly elevated. This is most severely the case for the dense reward which increases with the number of network nodes. 
The sparse positive reward is uniquely antagonised by the blue then red agent order as the only circumstances for incentive, a network with no compromised nodes, is dependent on the red agent not performing a basic attack. As this occurs with only a 10\% probability, sparse positive rewards are made particularly sparse by this configuration.



\vspace{-0ex}

\begin{table}[ht]
\caption{Results for agents trained using 3 different reward functions for both agent orders in the extended action space.}
\vspace{0.1in}
\small
\centering
\renewcommand{\arraystretch}{1.2} 
\begin{tabular}{l c c c c c c c}
    \toprule
    \multirow{2}{*}{\textbf{Agent Order}} & \multirow{2}{*}{\textbf{Network Size}} 
    & \multicolumn{6}{c}{\textbf{Reward Function Evaluation -- Action space: Extended}} \\  
    \cmidrule(lr){3-8} 
    
    & & \multicolumn{2}{c}{\textbf{Sparse Postive}} & \multicolumn{2}{c}{\textbf{Sparse Negative}} & \multicolumn{2}{c}{\textbf{Dense}} \\  
    \cmidrule(lr){3-4} \cmidrule(lr){5-6} \cmidrule(lr){7-8}
    
    & & \textbf{Eval Score}  & \textbf{$\bar{\text{DV}}$ (e-3)} 
    & \textbf{Eval Score}  & \textbf{$\bar{\text{DV}}$ (e-3)} 
    & \textbf{Eval Score}  & \textbf{$\bar{\text{DV}}$ (e-3)} \\
    
    \midrule
    
    \multirow{5}{*}{\textbf{Red then Blue}}  
    & 2  & -0.90  & 0     & -1.31    & 0      & -0.90     & 0       \\
    & 5  & -0.90  & 0     & -3.90    & 0      & -1.13     & 0.30  \\
    & 10 & -0.90  & 0     & -7.39    & 0      & -1.82     & 0.58  \\
    & 20 & -0.90  & 0     & -11.09   & 0.01   & -4.29     & 3.34  \\
    & 50 & -0.90  & 0     & -22.84   & 0      & -12.53    & 9.45  \\ 
    
    \midrule
    
    \multirow{5}{*}{\textbf{Blue then Red}}  
    & 2  & 0     & 0         & -0.89    & 0.26   & -0.68   & 0.60   \\
    & 5  & 0     & 0         & -3.06    & 0.31   & -1.03   & 0.67   \\
    & 10 & 0     & 0         & -6.84    & 0.47   & -1.48   & 0.84  \\
    & 20 & 0     & 0         & -9.95    & 0.17   & -3.30   & 4.67   \\
    & 50 & -1.36 & 0.94      & -24.33   & 0.01   & -8.42   & 4.90   \\ 
    
    \bottomrule
\end{tabular} \label{tab: decoy action space}
\end{table}


Finally, we evaluate blue agents with the extended action space which enables placing decoys on the network to proactively defend against attacks. The best average score remains $-0.9$ for red then blue agent order, but is $0$ for blue then red because the red agent can now be proactively prevented from corrupting a node at the end of a time step. For the same reason, the previously exacerbated sparsity of the sparse positive reward is also eliminated. As shown in Table~\ref{tab: decoy action space}, the sparse positive reward again outperforms the sparse negative and dense reward functions, achieving the optimal score on average in each network size up to 20 nodes and showing superior training reliability. Performance across all three reward functions is lowest at the largest network size of 50 nodes.



\begin{figure}[ht] 
    \centering 
    \includegraphics[width=0.75\textwidth]{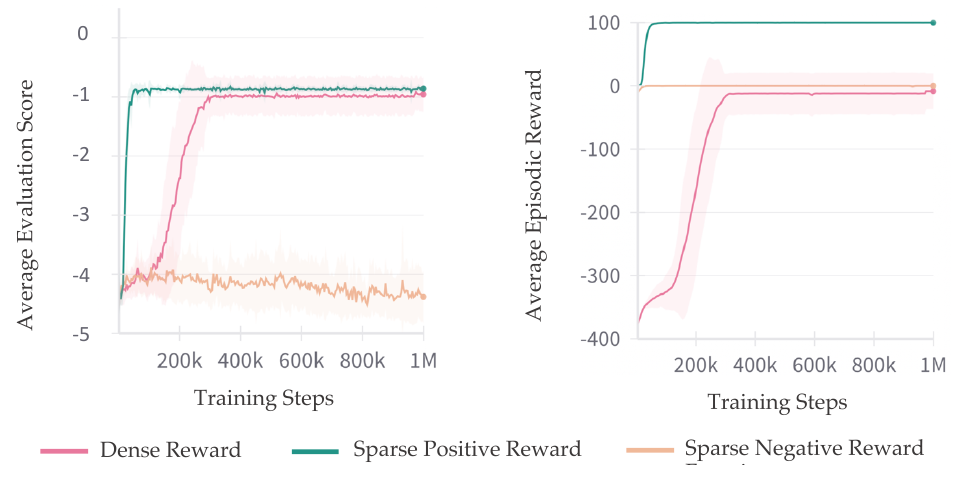} 
    \caption{The average episodic reward (left) and evaluation score (right) training curves for blue agents in a 5 node network using the basic action space and the red then blue agent order .}
    \label{fig:5 Nodes training and eval curves} 
\end{figure}

Figure \ref{fig:5 Nodes training and eval curves} compares the episodic rewards and evaluation score throughout training when the network size is 5, the agent order is red then blue, and agents use the basic action space. Despite offering comparatively infrequent incentives, the sparse positive reward shows both a steeper policy-improvement gradient and greater stability across time.




\vspace{-2ex}


\section{Conclusion}

In this work we evaluate whether sparse reward functions could enable more effective ACD agents to be trained. We introduce a ground truth evaluation score, and two types of sparse reward function, which when implemented in a popular cyber gym show that sparse rewards can provide both more effective ACD agents and increased training reliability. Furthermore, our results demonstrate the complex inter-relationship between reward structure, action space, and MDP definitions in the context of cyber defence.

\bibliography{References}

\begin{thebibliography}{1}

\bibitem{yawningtitan}
Alex Andrew, Sam Spillard, Joshua Collyer, and Neil Dhir.
\newblock Developing optimal causal cyber-defence agents via cyber security simulation.
\newblock In {\em Workshop on Machine Learning for Cybersecurity (ML4Cyber)}, 07 2022.

\bibitem{cage_cyborg_2023}
Cyber operations research gym.
\newblock \url{https://github.com/cage-challenge/CybORG}, 2022.
\newblock Created by Maxwell Standen, David Bowman, Olivia Naish, and others.

\bibitem{bates2023reward}
Elizabeth Bates, Vasilios Mavroudis, and Chris Hicks.
\newblock Reward shaping for happier autonomous cyber security agents.
\newblock In {\em Proceedings of the 16th ACM Workshop on Artificial Intelligence and Security}, pages 221--232, 2023.

\bibitem{riedmiller18a}
Martin Riedmiller, Roland Hafner, Thomas Lampe, Michael Neunert, Jonas Degrave, Tom van~de Wiele, Vlad Mnih, Nicolas Heess, and Jost~Tobias Springenberg.
\newblock Learning by playing solving sparse reward tasks from scratch.
\newblock In {\em Proceedings of the 35th International Conference on Machine Learning}, Proceedings of Machine Learning Research, 2018.

\bibitem{Chan2020Measuring}
Stephanie~C.Y. Chan, Samuel Fishman, Anoop Korattikara, John Canny, and Sergio Guadarrama.
\newblock Measuring the reliability of reinforcement learning algorithms.
\newblock In {\em International Conference on Learning Representations}, 2020.

\bibitem{kiely2023autonomous}
Mitchell Kiely, David Bowman, Maxwell Standen, and Christopher Moir.
\newblock On autonomous agents in a cyber defence environment.
\newblock {\em arXiv:2309.07388}, 2023.

\end{thebibliography}

\end{document}